# Coupling Machine Learning with Ontology for Robotics Applications

Osama F. Zaki[1,2]

[1]Robotaar, Livingston, EH54 6DD, Scotland, UK, osama.zaki@robotaar.com
[2]Sinai University, Arish, Sinai, Egypt, osama.farouk@su.edu.eg

**Abstract**

*In this paper I present a practical approach for coupling machine learning (ML) algorithms with knowledge bases (KB) ontology formalism. The lack of availability of prior knowledge in dynamic scenarios is without doubt a major barrier for scalable machine intelligence. My view of the interaction between the two tiers intelligence is based on the idea that when knowledge is not readily available at the knowledge base tier, more knowledge can be extracted from the other tier, which has access to trained models from machine learning algorithms. To analyse this hypothesis, I create two experiments based on different datasets, which are related directly to risk-awareness of autonomous systems, analysed by different machine learning algorithms (namely; multi-layer feedforward backpropagation, Naive Bayes, and J48 decision tree). My analysis shows that the two-tiers intelligence approach for coupling ML and KB is computationally valid and the time complexity of the algorithms during the robot mission is linear with the size of the data and knowledge.*

1. Introduction

Trust in the reliability and resilience of autonomous systems is paramount to their continued growth, as well as their safe and effective utilization [32][9][7]. Hauser [13] reported the need for intelligent autonomous systems – based on AI and ML – operating in real-world conditions to radically improve their resilience and capability to recover from damage. Rich [23] expressed the view that there is a prospect for AI and ML to solve many of those problems. Cave and Dihal [5] claimed that a balanced view of intelligent systems by understanding the positive and negative merits will have impact in the way they are deployed, applied, and regulated in real-world environments.

AI and robotics researchers have applied ontology as a knowledge-based scheme, within a system to support robotics autonomy, such as SMERobotics [20], KnowRob 2.0 [30], CARESSES [4], open-EASE [3], ORO [27][17], SIARAS [12]. They covered a spectrum of cognitive functions, which according to the classification made by [16] and [28] are recognition and categorization, decision making and choice, perception and situation assessment, prediction and monitoring, problem solving and planning, reasoning and belief maintenance, execution and action, interaction, and communication, and remembering, reflection, and learning. The ontology scope of these prior works varies, and it depends on the functionalities of the target robotic system, i.e. concepts that were modelled in the ontology are related to: object names, environment, affordance, action and task, activity and behaviour, plan and method, capability and skill, hardware components, software components, interaction, and communication [18][33]. Primary motivation for the use of ontologies within robotics is that these knowledge-based approaches offer an expandable and adaptable approach for capturing the semantic features of model robot cognitive capabilities. Furthermore, when considering a fleet distribution of robotic platforms, or swarms, the ontology provides a cyber-physical interface to cloud, web-based

[Type here]

service robots, such as RoboEarth [29] and openEASE [2] that enable robots to collect and share knowledge of missions. This knowledge enabled architecture provides a means of sharing knowledge via the ontology, between different robots, and between different subsystems of a single robot's control system in a machine understandable and consistent presentation. Therefore, attempts have been made to create CORA (Core Ontology for Robotics and Automation), which was developed in the context of the IEEE ORA (Ontologies Robotics and Automation) working group. However, creating a complete framework, which is a highly complex task, was outside the scope of the ORA working group initiative [22]. The 1872–2015 IEEE Standard Ontologies for Robotics and Automation defines an overall ontology which includes key terms as well as their definitions, attributes, constraints, and relationships. Sub-parts of this standard include a linguistic framework, generic concepts (an upper ontology), a methodology to add new concepts, and sub-domain ontologies [26]. The core ontology was utilised in some projects such as [11][6]. In the PANDORA framework [15], ontologies are used as a way for the robot to organise the knowledge about the world, not just in geometric concepts, but by attaching a semantic label. A diagnostic module has been developed to detect a thruster failure, with a link between the degrees of freedom of the vehicle and the executable actions. In [31], a modelling paradigm based on ontology for online diagnostics and prognostics for autonomous systems is presented. During the work in [31], there were some areas where knowledge for the purpose of safety and reliability is not readily available. This has been a main motive to couple ML algorithms with the ontology. For example, it could be very useful to determine the health of the battery before and during the robot mission. This piece of knowledge is not readily available. Hence, we need to extract knowledge, via ML algorithms, from the live data collected from different sensors before populating it into the Ontology. For example, in [24], ML is used to determine the battery state of health (SOH) for ultimately safeguarding asset integrity. Two parametric and two non-parametric algorithms are used to estimate battery SOH. The SOH value can be predicted by the ML algorithm and then populated into the ontology.

The two-tiers intelligence concept will enhance the learning and knowledge sharing process in a setup that heavily relies on some sort of symbiotic relationships between its parts and the human operator [36]. It will also assist in explaining the behaviour of the robot. Philosophically, human has more than one tier of intelligence, while animals have one-tier intelligence, which is the intrinsic and the static know-how. The harmony between the two tiers can be viewed from different angles, however they complement each other, and both are mandatory for human intelligence and hence machine intelligence. There are many applications that can benefit from the two-tiers intelligence concept in robotics systems, their reliabilities, and several other non-robotics applications.

Although the work in this paper is a contribution to the ongoing research of integrating the two existing thoughts for modelling cognition and intelligence in humans [34][35], my goal is not just to investigate the neuro-symbolic combination, but rather is to investigate a wider range of inductive machine learning algorithms within the context of cognitive robotics. The following section highlights the theory that backups the two-tier intelligence concept, Section 2. Experiments and results are shown in Section 3. The computational complexity is analysed in Section 4.

2. The ML-KB Coupling Mechanism

The term Knowledge Discovery in Databases (KDD) defined steps to extract knowledge from data in the context of large databases [8]. It defines five stages to discover knowledge from raw data in a database into a knowledge base: Selection, Processing, Transformation, Data Mining, Interpretation/Evaluation.

[Type here]

Therefore, the approach of coupling in this paper has two folds; firstly, to use the data collected during a robot's task (a robot mission), i.e. inspection, to discover new relationships or associations between different elements, secondly, to interpret those relationships into existing knowledgebase by applying a semi-automated or fully automated process. This would result in a ML trained models supporting ontology-based decision-making for robots when relationships between different elements are unclear.

To use the data collected by the robot, it must go through pre- and post-processes, after that, the three main steps are: 1) choosing the most suitable ML algorithm, 2) evaluation and interpretation of the trained model, and 3) encoding the knowledge into the symbolic system (knowledge base). This means that the robot has learned new knowledge after performing his first mission which enhance its performance for the following missions. One of the main challenges is that the trained models (or learned patterns) which are the output of ML algorithms, in most, produce models which are unreadable by humans (i.e., binary coded), unlike J48. The final goal of our research is to carry this ontology learning process at online and without human interactions (fully automated), if possible.

### 3. Experiments and Results

3.1 Experiment One

The first experiment is selected to prove the two-tiers concept, but it is also related to kind of inspection that is carried by a Husky (a robot platform) which is identifying types of materials. The training dataset in this experiment came from the glass database from the USA Forensic Science Service [14]. Six types of glass are defined in terms of their oxide content (i.e., Na, Fe, K, etc). The dataset has 214 instances, 9 attributes, and 7 classes, as shown in **Table 1**.

**Table 1: Dataset parameters and classes**

| |
|---|
| 1. Id number: 1 to 214 |
| 2. RI: refractive index |
| 3. Na: Sodium (unit measurement: weight percent in corresponding oxide, as are attributes 4-10) |
| 4. Mg: Magnesium |
| 5. Al: Aluminum |
| 6. Si: Silicon |
| 7. K: Potassium |
| 8. Ca: Calcium |
| 9. Ba: Barium |
| 10. Fe: Iron |
| 11. Type of glass: (class attribute) |
| -- 1 building_windows_float_processed |
| -- 2 building_windows_non_float_processed |
| -- 3 vehicle_windows_float_processed |
| -- 4 vehicle_windows_non_float_processed (none in this database) |
| -- 5 containers |
| -- 6 tableware |
| -- 7 headlamps |

Five ML algorithms were applied to the dataset, using the n-fold test option cross validation procedure which is used to estimate the skill of the model on new data, with the value of 'n' equal to 10. The ML algorithms are: J48, Naïve Bayes, SVM (SMO), Logistic Regression, and Multilayer Perceptron. Performance is measured with reference to the following parameters: correctly classified instances (CCI), incorrectly classified instances (ICS), Kappa statistic (KS), mean absolute error (MAE), root mean squared error (RMSR), relative absolute error (RAE), and root relative squared error (RRSE). **Table 2** summaries the results from applying the five ML algorithms.

[Type here]

Table 2: Performance and Evaluation of different ML algorithms

| Algorithms/Parameters | CCI % | ICS % | KS | MAE | RMSE | RAE % | RRSE % |
|---|---|---|---|---|---|---|---|
| **J48** | 66.82 | 33.18 | 0.55 | 0.10 | 0.29 | 48.45 | 89.27 |
| **Naïve Bayes** | 48.60 | 51.40 | 0.32 | 0.15 | 0.34 | 72.78 | 104.74 |
| **SVM (SMO)** | 56.07 | 43.93 | 0.36 | 0.21 | 0.32 | 100.94 | 97.56 |
| **Logistic-Regression** | 64.49 | 35.51 | 0.51 | 0.12 | 0.27 | 57.09 | 84.57 |
| **Multilayer-Perceptron** | 67.76 | 32.24 | 0.55 | 0.11 | 0.26 | 52.59 | 80.96 |

Next, I demonstrate the two-tiers intelligence concept with three selected ML algorithms: J48, Naive Bayes and Multilayer Perceptron. Scikit [19] is used. Weka [24] is also possible framework for ML with a user-friendly interface and it was also tired. **Table 3** shows a portion of the output pruned tree of J48 as an example, which the easiest to interpret.

Table 3: A portion of the output J48 pruned tree

```
Ba ≤ 0.27
|   Mg ≤ 2.41
|   |   K <= 0.03
|   |   |   Na <= 13.75: build wind non-float (3.0)
|   |   |   Na > 13.75: tableware (9.0)
|   |   K > 0.03
|   |   |   Na <= 13.49
|   |   |   |   RI <= 1.5241: containers (13.0/1.0)
|   |   |   |   RI > 1.5241: build wind non-float (3.0)
|   |   |   Na > 13.49: build wind non-float (7.0/1.0)
```

In the knowledge base side, the formalism used for modelling is description logic (DL) which is a subset of first order predicate logic, i.e., an ontology-based representation [21]. The Ontology Web Language (OWL) is a suitable platform for experiments and demonstrating the proof of concept. The interpretation of the pruned tree is done semi-automatically (scripting and API). SWRL rules are created based on the information provided by the tree and the knowledge base is then extended, Step 3. The modelling process is described in **Table 4** while **Figure** visualises the outcome of the modelling process.

Table 4: The modelling process of the ontology

- Let $T$ represents the set of glass types such that $t$ is a type of glass:
    $T = \{t_1, t_2, ..., t_n\}$
- Let $A$ represents the set of glass attributes such that $a$ is an attribute (glass data) of the glass:
    $A = \{a_1, a_2, ..., a_n\}$
- Let $R$ represent a set of semantic rules such that r is a SWRL rule:
    $R = \{r_1, r_2, ..., r_3\}$
- Let $I$ represents a set of individuals such thaI $i$ relates a type of glass $t$ with distinguished glass attributes $A'$, $A'$ is a subset of $A$, and $i \equiv (t, A')$:
    $I = \{(t, A') \mid t \in T, A' \subset A\}$
- The data property *has_value* is of type float which relates values to attributes.
- Sets and their elements are organised in hierarchal style using the *subClassOf* relationship.

[Type here]

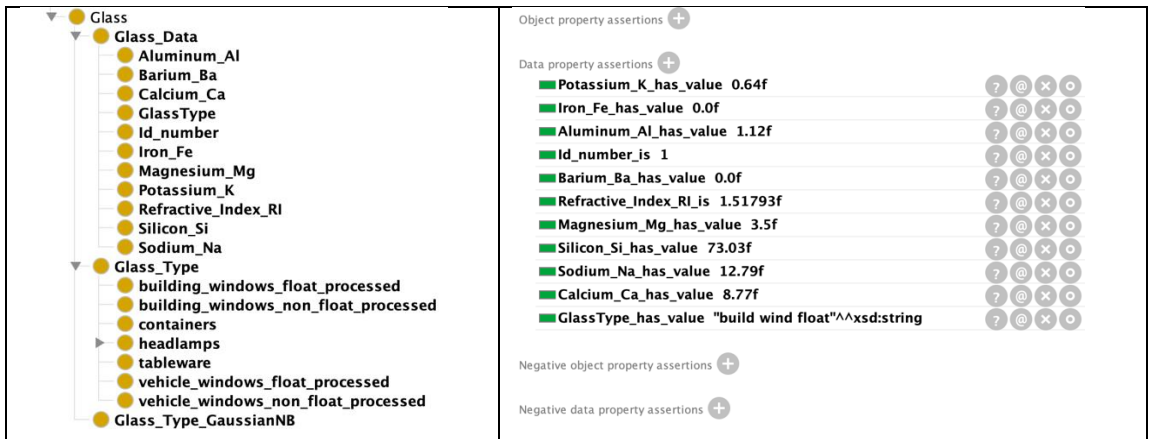

**Figure 1: A schematic view of the ontology concepts and data properties**

In this experiment, the appropriate ML algorithms are applied offline to the already exist dataset. Then the trained model is embedded into the online system, the output of the trained model of the unlabled data is parsed by semi-automated process (scripting, API and human intervention), the new extracted knowldge is inserted (scripting, API and human intervention) into the knowledge base while the ontology reasoning algorithm is running.**Table 5** shows the top-level functions of the ontology tier algorithm in Python-like pseudocode.

**Table 5: Top level functions of the ontology tier algorithm**

```
def classify_glass():
    data = scan_glass_data()
    onto = populate_into_ontology(data)
    individuals = apply_reasoner(onto)
    glass_type = identify_glasses (individuals)
return glass_type
```

It is also possible in the ROS (Robotics Operating System) environment to have both tiers applied online. In such scenario, data is streamed from sensors and published into a topic. Another ROS node prepares the dataset and then applies a chosen learning algorithm. This means that the trained model by then is already embedded into the system. The procedures are shown in **Figure** .

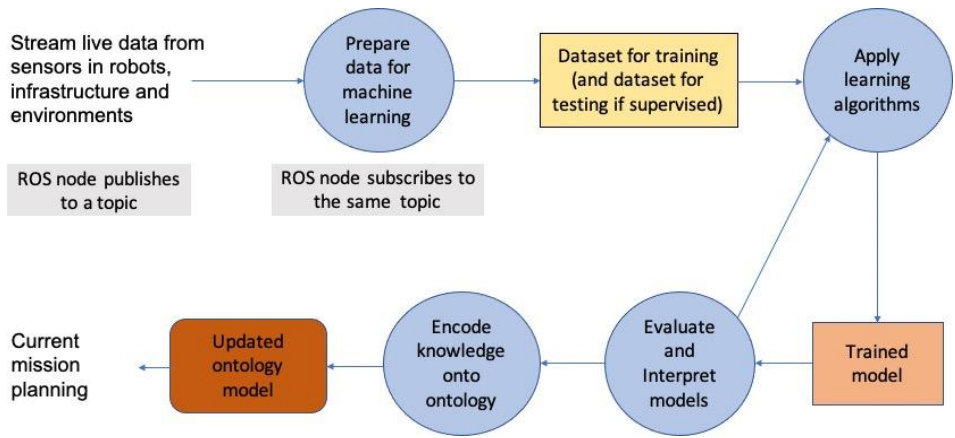

**Figure 2: Generic procedures for online two-tiers intelligence concept**

[Type here]

3.2 Experiment Two

In the second experiment, the robot was commanded to autonomously navigate inside an enclosed test arena. Five test cases were designed to mimic real-world failure scenarios by purposely manipulating either Jackal's (a robot platform) sensor data or hardware to interfere with its autonomous navigation. The five test cases are: 1) Lidar interruption, 2) IMU interruption, 3) Odometry drift, 4) Deflated tyre, and 5) Unseen obstacle.

The first test case considers hardware failure that would affect the Lidar. Such failure could be associated with, for example, sensor malfunction, cable breakage, or an open circuit. This experiment examines the impact of the absence of incorrect or incomplete data on the AMCL module used as localisation for the Move base navigation stack. The second test case considers the impact of a malfunctioning IMU and the impact of that on the EKF (Extended Kalman Filter) pose estimation module in the Jackal control package. The third and fourth test cases, odometry drift and deflated tyre, consider the impact of the hardware failure in the wheel-encoder and the EKF pose estimation, Figure 5. Multiple weights, totalling approximately 5 kg, were added to the front right side of the robot to increase the load on the tyre, Figure 3. The last test case examines an external opposing force against the robot's forward direction affecting the robot's localisation. The external opposing force can be caused by an object that is outside the lidar field of view of the robot used in these test cases, Figures 4.

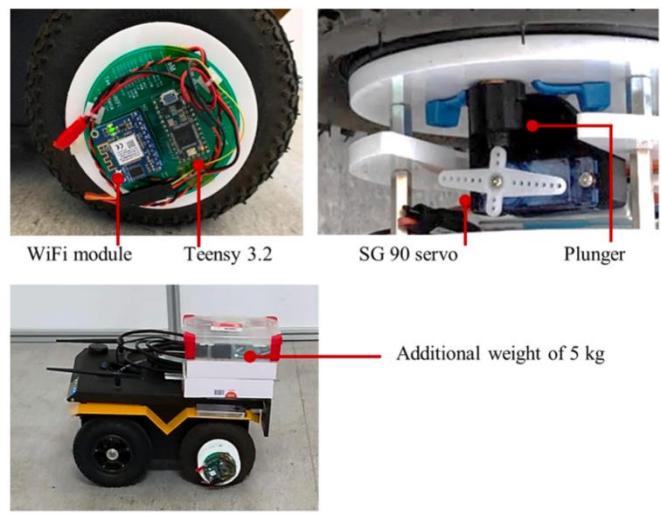

Figure 3: Remote triggered tyre deflating device

[Type here]

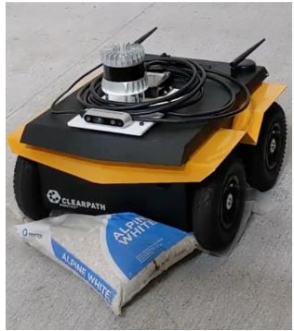

**Figure 4: Jackal stuck with an unseen object.**

The robot is equipped with an Ouster OS1 lidar and an Intel RealSense D453i depth camera installed on the top of the robot. In addition, a tyre deflating device was installed on the front right wheel. The tyre deflating device was designed and developed, by the team at Manchester University, to reduce tyre pressure in a controlled manner to simulate a flat tyre. The device consists of a Schrader valve pin plunger which can be activated by an SG 90 servo. Once activated, the plunger pushes the metal pin in the Schrader valve to allow air to escape from the tyre. The device was controlled by a Teensy 3.2 which communicates wirelessly with the base station PC via *rosserial* and an Adafruit ATWINC1500 WiFi Module, using methodology presented in [23]. A 200 mAh LiPo battery was used to power the device independently. Zip ties were used to fix the device to the robot's front right wheel, Figure 5.

For all five test cases, the robot was commanded to autonomously navigate continuously along a 1.6-by-1.6 m square route defined by four waypoints (WP 1, 2, 3 and 4). A base station PC was remotely connected to Jackal for initiating the robot, injecting programmatic data manipulation, and recording sensor and diagnostic data, Figure 6.

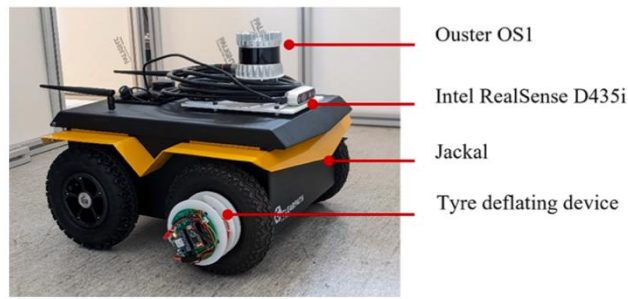

**Figure 5: Jackal robot setup**

[Type here]

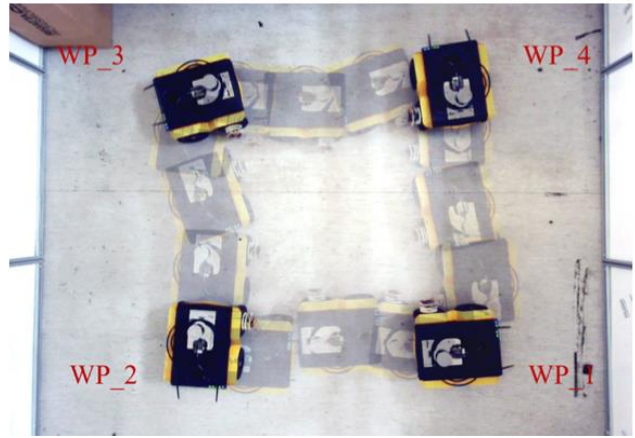

**Figure 6: Jackal navigates around a square route defined by four waypoints.**

The robot was first commanded to complete two full laps around the test arena for all five test cases, with normal tyre pressure (approximately 30 PSI) on all four wheels, to collect baseline data. At the end of the second lap, faults were remotely triggered on the base station PC, except for the unseen obstacle experiment, where a heavy bag of stone chippings was added to the midway point between WP 4 and WP 1.

In Jackal's semantic model, hardware parts (as concepts) of Jackal system are represented as classes (sets) in a hierarchical manner (taxonomies). Classes can be made disjoint. Each set has zero or more individuals. Data (values) can be assigned to individuals. The relationships between parts of the system are declared as facts between the classes, and then the individuals (objects) bind during the reasoning process. The relationships between classes/sets are represented as a relation between Domain and Range. Other characteristics can be added to the relation so that the relation can be made functional, transitive, symmetric, reflexive, and their inverse. Basic semantics of the relations can be expressed directly using the ontology formalism and others use SWRL. After the model of the systems is built, a reasoner is used to check the consistency of the model and rules are applied. Further queries can be made using the Description Logic (DL) interface or SPARQL queries. The diagnosis automaton (or diagnoser) is the automaton which defines the semantics for the states and arcs for each part of the system [31]. Figure 7 shows (to the left) the main classes for the hardware setup; chassis, external devices, and Jackal internal system. The figure also shows (to the right) the SW and ROS packages, topics, and messages.

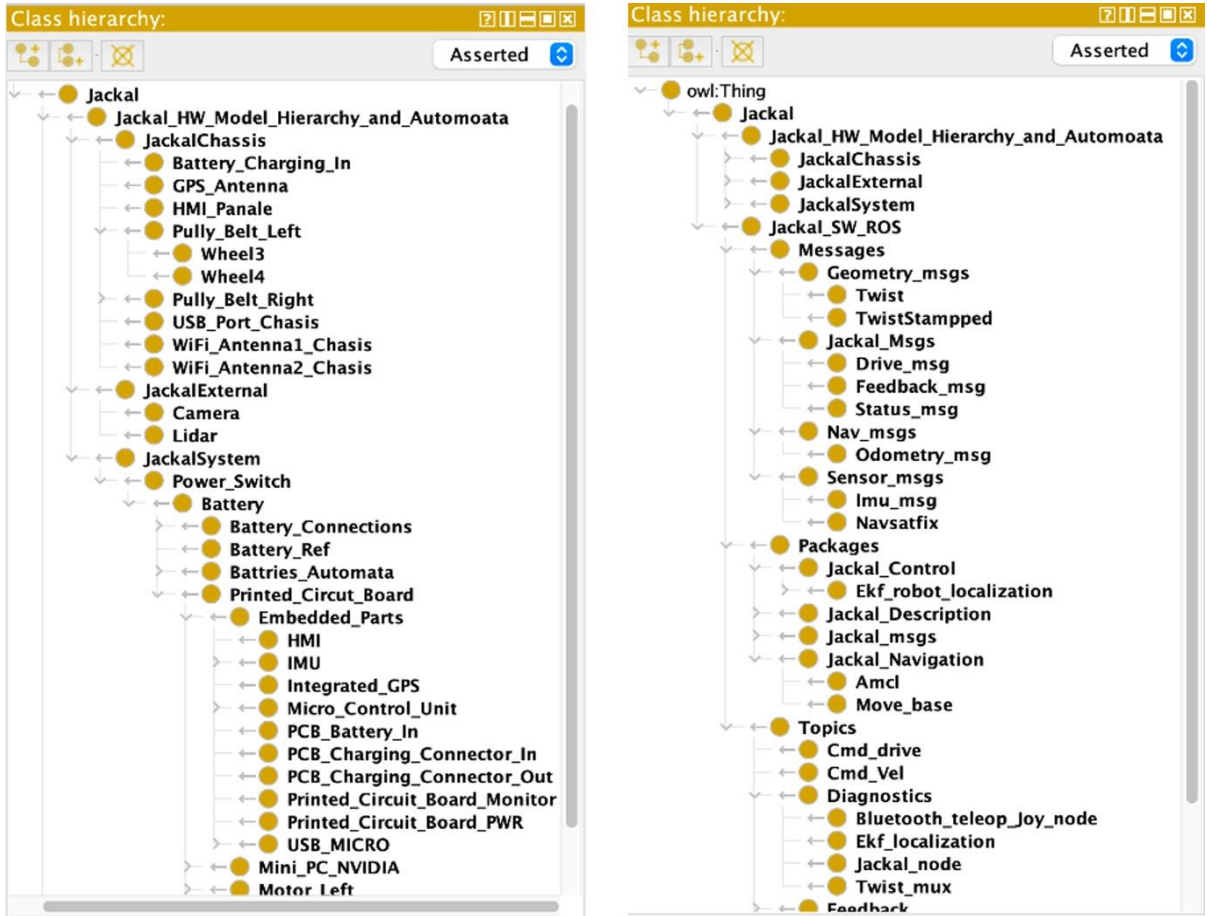

**Figure 7: The main classes for HW and SW and ROS packages.**

Jackal's model is based on the *hierarchical-relationship* model that was developed for Husky robot in [31]. Two models are created, the semantic of the hierarchical relationship in the first model is *'is-type-of'* relationship, while in the second model it is *'is-linked-to'* or *'is-connected-to'* relationships, for example, *'x is-connected-to y'*. This means the system to be diagnosed has two models, the first one is more generic than the second and it can be adopted easily by other systems. The second model is the concern of this paper since the first model is just representing the taxonomies of the system components, for example, *'temp-sensor is-type- of sensor'*. The other relationships are semantically declared in the second model between two different components and between two specific states in different automata. For example, *'state2 x causes-high-temp-to state4 y'*. The relationship *'causes-high-temp-to'* between *state2* and *state4* is declared in the model, abstractly, without binding to any specific states or components, while the relationship *'may- causes-high-temp-to'* between *state2* and *state4* is defined between members of *state2* (the domain) and all members of *state4* (the range). The relationship *'may-causes-high- temp-to'* is inferred (or asserted) by the reasoner, while the relationship *'causes-high-temp-to'* is inferred by the semantic rules. Figure 8 shows the object properties and annotations sections, and Figure 9 shows the data properties and the individual sections of the ontology.

[Type here]

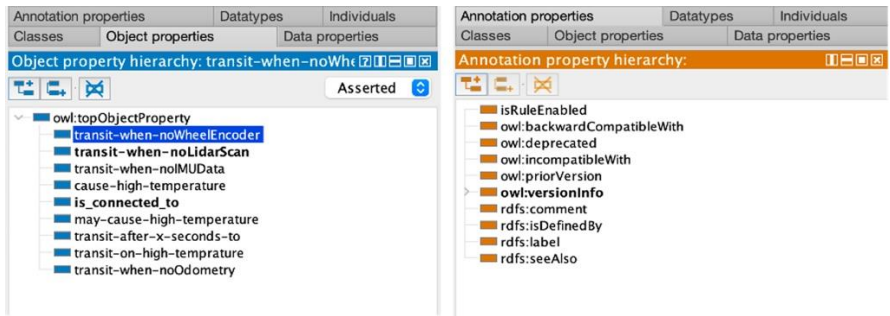

**Figure 8: Object properties and annotation properties of the ontology**

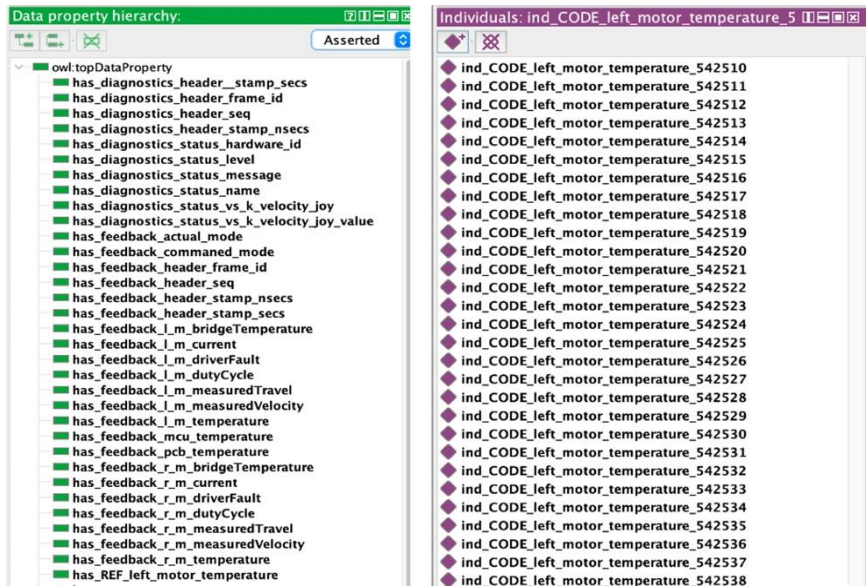

**Figure 9: Data properties and individuals of the ontology**

The Multilayer-Perceptron algorithm was selected and applied, and data were prepared accordingly and encoded. This task was very complex because within a 5-minutes mission each case produced about 200MB of data distributed over 82 CSV files. The number of features available are too many and they are distributed over those 82 CSV files. Therefore, only the following 14 CSV files were selected:

cmd_drive (number of rows: 9296)
cmd_vel  (1721)
diagnostics (50342)
feedback (3691)
jackal_velocity_controller-cmd_vel (1721)
jackal_velocity_controller-odom (9312)
move_base-current_goal (25)
move_base-feedback (1719)
move_base-result (25)
move_base-status (8185)
odometry-filtered (9338)
status (188)

Then, he number of features/files are reduced if they don't match closely. The final selected 3 CSV files were:

[Type here]

cmd_drive (rows: 9296)
move_base-status (8185)
odometry-filtered (9338)

Those 3 CSV files were selected from each of the five test-cases, plus the two base cases (normal behaviour with and without the weight on the robot), this resulted in 7 datasets:

baseline_jackal,
weightedbaseline_jackal,
flatyre_jackal,
imuIntercept_jackal,
lidarIntercept_jackal,
odomDrift_jackal,
unseenObstacle_jackal,

Each dataset was then divided into a training set (80% of total rows) and a testing set (20% of total rows). Two output labels were then set: "Normal", "UpNormalXXX", where "XXX" refers to a different dataset. All data for training are then grouped into one file, and all data for testing are also grouped into one file. At the end, I have one file of size (9000 rows * 80% * 7) for training and another file of size (9000 rows * 20% * 7) for testing. The results of ML were interpreted semi-automatically with the help of the expert. The new knowledge was encoded using a program but aided by human-interaction and manipulation, SWRL rules were generated and inserted into the ontology to diagnose and to identify the cause of the problem and predict the fault/failure in advance. Once the new knowledge is encoded, the Pellet reasoner is run to check consistency of the knowledgebase and then making the decision. A SWRL rule might look like this:

> has REF left motor temperature(ind REF left motor temperature, ?valueREF) has feedback l m temperature(?x, ?valueLive) swrlb:greaterThan(?valueLive, ?valueREF) -¿ Motor Left Resting(?x)

Other rules can achieve the following: 1. direct the robot to return to an emergency waypoint relying on IMU and GPS data 2. indicate that there is only an intermittent fault, and the robot can continue with the mission. 3. determine whether the Lidar is faulty or not. 4. the robot can continue the mission even with the interrupted data from the IMU. 5. the drift in the IMU data is detected and directs the robot to change its plan.

4. Performance and Overall Analysis

For the ontology, two complexity and scalability tests are performed:

1. The ratio in size between the theoretical size (raw data to be populated into the ontology) and the actual size of the ontology after raw data was populated into it. This test indicates the complexity of the space required (storage or working memory).
2. The time taken by the reasoning process versus the size of the ontology when it is loaded into working memory. This test indicates the complexity of time required by the reasoning process.

The ontology experiment in [31] showed that the space required by the ontology is about 25 times more than the size of the raw data on average, and there is a linear relationship between the two variables. Interestingly, this is also true for our case presented in this paper. When the raw data is 14KB the ontology is about 343KB (24.5 times the raw data). The time taken by the reasoner (the reasoning

[Type here]

process) is almost linear with respect to the size of the ontology when loaded into working memory. For example, when the size of the ontology is 4MB the computational time taken is 30s to achieve run-time diagnostics. For ontology of size 0.343MB the computational time taken is about 2 seconds.

From the ML side, the time taken for the glass dataset which of size 14KB (like the size of data used in the ontology experiment in [31]) for different algorithms is shown in **Table 6**. This shows that the most time taken is by the multilayer-perceptron is 0.4s.

Table 6: Performance and Evaluation of different ML algorithms

| Algorithms/Parameters | Time (s) |
|---|---|
| **J48** | 0.04 |
| **Naïve Bayes** | 0.001 |
| **SVM (SMO)** | 0.2 |
| **Logistic-Regression** | 0.19 |
| **Multilayer-Perceptron** | 0.4 |

This means the total time taken by the system reading the logging data from sensors, applying J48, for example, to extract knowledge, feed it into the ontology, and finally perform reasoning is achieved in less than 2.1s. With such small size data, it is reasonable to perform the training online during the missions. However, for experiment Two, applying the Multilayer-Perceptron on a training dataset of size 240MB took nearly 7200s (on 2.6x6 GHz processor, 16MB of RAM). This clearly means, it is not reasonable to apply the ML algorithm online, i.e., during the mission. Adding to this, the reasoning time is significantly longer with such large data.

**Conclusion**

As other AI researchers who have recognised the potential and advantages of integrating deductive reasoning and inducive learning together, this paper has demonstrated a practical example of coupling three ML algorithms with a knowledge base in a real-life application in the robotics domain. There are still challenges; to make this coupling fully automatic without the human expert intervention, to devise an architecture for neuro-symbolic integration, and - last but not least - addressing scalability issues. Despite the challenges, undoubtedly, I believe it is an innovative contribution and it would be of great benefit for other researchers. I further believe that to the best of my knowledge no similar work has been reported to combine ML algorithms with KB in the robotics domain.

**Acknowledgment**

I would like to thank Zhengyi Jiang, Andrew West, and Simon Watson, Centre for Robotics and AI, at Manchester University for their help while carrying out the experiments. Also, thanks go to David Flynn, Glasgow University for his general comments on the research. Special thanks to Matthew Dunnigan, at Heriot Watt University, for editing early versions of the manuscript and his continuous support. Finally, thanks go to my student Taha Senoosy who helped in software implementation and testing.

[Type here]

[Type here]

[Type here]